\documentclass[times, review, 10pt]{elsarticle}
\usepackage{amssymb}
\usepackage{setspace}
\usepackage{amsmath}
\usepackage{graphicx}
\usepackage{caption}
\usepackage{float}
\usepackage{booktabs}
\usepackage{multirow}
\usepackage{makecell}
\usepackage{bbding}
\usepackage[normalem]{ulem}
\useunder{\uline}{\ul}{}
\usepackage{adjustbox}
\usepackage[section]{placeins}
\usepackage[colorlinks=true, allcolors=blue]{hyperref}
\usepackage{lineno,hyperref}
\usepackage{threeparttable}
\usepackage{ulem}
\usepackage{color}
\usepackage{tabularx}
\journal{Nuclear Physics B}
\usepackage{cleveref}
\Crefname{Figure}{Fig}{Figures}

\begin{document}
\doublespacing
\begin{frontmatter}
\title{Progressive Pixel-Neighborhood Deformable Cross-Attention for Multispectral Object Detection}

\author[a]{Tian Qiu}
\author[a]{Jifeng~Shen\corref{mycorrespondingauthor}}
\cortext[mycorrespondingauthor]{Corresponding author}
\ead{shenjifeng@ujs.edu.cn}

\author[b]{Xin~Zuo}
\address[a]{School of Electrical and Information Engineering, Jiangsu University, Zhenjiang, 212013, China}
\address[b]{School of Computer Science and Engineering, Jiangsu University of Science and Technology, Zhenjiang, 212003, China}

\begin{abstract}
Effective cross-modal feature alignment and interaction are central challenges in multispectral object detection. Although global cross-attention provides strong long-range modeling ability, its quadratic complexity with respect to feature size limits deployment on resource-constrained platforms. We therefore propose Progressive Pixel-Neighborhood Deformable Cross-Attention for multispectral feature fusion, termed PNAFusion. The proposed framework is motivated by two observations: weak misalignment between visible and thermal images is usually concentrated around local neighborhoods, and semantic correspondence across modalities often follows non-linear spatial mappings that fixed receptive fields cannot model well. To address these issues, PNAFusion incorporates local spatial priors into its architectural design to concentrate feature interaction and alignment on the most relevant neighborhoods. Specifically, a Pixel-Neighborhood Cross-Attention (PNCA) module is introduced to avoid redundant global feature matching and suppress background noise. Meanwhile, an Adaptive Deformable Alignment (ADA) module captures non-linear spatial correspondences through learned pixel-wise offsets. These components are further integrated through an iterative feedback mechanism to progressively refine cross-modal feature alignment. Experiments on FLIR, M3FD, and DroneVehicle show that PNAFusion achieves 84.2, 90.5, and 85.5 mAP@0.5, respectively, under the YOLOv5 detector, and further reaches 86.8 mAP@0.5 on FLIR and 90.8 mAP@0.5 on M3FD when transferred to Co-DETR. Efficiency analysis indicates that PNAFusion reduces allocated GPU memory by 33.0\% compared with ICAFusion and reduces theoretical FLOPs from 194.8 G to 156.4 G, although the deformable sampling and iterative refinement introduce additional latency. These results demonstrate that PNAFusion provides a practical accuracy--memory trade-off for weakly aligned multispectral object detection.
Our code will be available at https://github.com/DanielQiuTian/PNAFusion.
\end{abstract}

\end{frontmatter}


\section{Introduction}

In safety-critical domains such as autonomous driving, all-weather security surveillance, and~UAV-based remote sensing~\cite{ref1,ref2}, robust object perception constitutes a fundamental prerequisite for system safety. Although~current RGB (visible-spectrum) detection algorithms have achieved remarkable performance under ideal illumination conditions, they remain highly dependent on ambient lighting. Under~extreme conditions, such as nighttime, intense glare, and~dense fog, thermal infrared (TIR) sensors can  provide reliable feature cues by capturing the thermal radiation emitted by objects. However, infrared images generally suffer from low resolution, missing texture details~\cite{ref3}, and~low signal-to-noise ratio. Therefore, how to effectively integrate visible-spectrum information with infrared thermal radiation has become a challenging problem in multispectral image perception. Multispectral object detection, which combines the texture information of visible (VIS) images with the thermal radiation characteristics of infrared (TIR) images, has substantial application potential. Nevertheless, significant difficulties and bottlenecks remain in achieving both efficient deep cross-modal feature interaction and high \mbox{computational~efficiency.}

The evolution of multispectral vision models has seen a paradigm shift from traditional local modeling to interactive and, more recently, global modeling strategies.  Early studies mainly relied on convolutional neural networks (CNNs) and various fusion operators to integrate information through feature-level stacking. However, due to the limited local receptive field of convolution kernels, such methods are inherently insufficient for long-range interaction modeling~\cite{ref4}. In~recent years, Transformers (ViTs), owing to their excellent long-range modeling capability, have demonstrated great potential in multispectral tasks~\cite{ref5,ref6}. ICAFusion~\cite{ref7} further introduced an iterative cross-attention mechanism, which significantly enhanced semantic consistency across modalities through multi-stage interaction. 
Although existing fusion strategies have achieved notable progress in exploiting modal complementarity through deep interaction, they still face two major challenges. First, Transformer-based architectures represented by ICAFusion leverage global cross-attention to model long-range dependencies, but~the computational complexity of standard cross-attention grows quadratically with image resolution, i.e.,~$\mathcal{O}((HW)^2)$. This imposes substantial memory overhead and may limit deployment on memory-constrained embedded platforms. Second, visible and thermal infrared images often exhibit weak spatial misalignment due to physical displacement between sensors, calibration errors, different imaging mechanisms, or~non-identical fields of view. Conventional fusion methods usually assume strict pixel-level registration; when this assumption is violated, feature ghosting, blurred object boundaries, and~localization bias may occur. Although~global attention can perform full-image matching, such many-to-many matching lacks explicit local deformation constraints and is easily distracted by background regions that are semantically irrelevant to the target. Therefore, an~effective multispectral detector should not only correct content-dependent local spatial offsets, but~also avoid unnecessary global cross-modal matching. To~support this design motivation more explicitly, we rely on both prior studies and empirical observations. In~dual-sensor multispectral systems, geometric deviations are often caused by limited baseline displacement and calibration residuals, which tend to produce local rather than arbitrary global misalignment. In~addition, our feature-response visualization  
shows that the ghosting artifacts of baseline fusion models are mainly distributed around object boundaries and their adjacent neighborhoods. This observation motivates us to constrain deformable sampling within a pixel neighborhood instead of searching across the entire image. Meanwhile, global cross-attention may introduce diffuse background activation, especially in cluttered traffic or aerial scenes. Such behavior motivates the use of local cross-modal attention with a restricted receptive field. Based on these considerations, we revisit the design of fusion Transformers and seek to answer two key~questions. 

\textbf{{How can robust cross-modal feature alignment be achieved under nonlinear spatial offsets?}} 
The accuracy of multispectral object detection largely depends on the spatial consistency between visible and infrared modalities. Conventional fusion frameworks usually assume strict pixel-level alignment, whereas in practice, nonlinear offsets caused by sensor parallax can substantially degrade localization accuracy. Different from previous alignment methods that rely on fixed coordinate transformations, this paper proposes an improved iterative deformable neighborhood adaptive alignment cross-attention fusion framework, termed PNAFusion. Inspired by content-aware offset prediction, the~proposed module is able to dynamically correct nonlinear displacements. By~adaptively adjusting sampling positions based on local semantic correlations, it ensures that complementary information is aggregated from accurately corresponding cross-modal~regions.

\textbf{{How can computational efficiency be balanced while preserving deep local interaction?}} Although global cross-attention mechanisms represented by previous ICAFusion demonstrate strong long-range modeling capability, their computational cost grows quadratically with image resolution, i.e.,~$O(H^2W^2)$. In~addition, global interaction inevitably introduces background noise, thereby weakening the discriminative representation of salient targets~\cite{ref10}. We observe that cross-modal complementarity is usually concentrated around the target and its adjacent spatial neighborhood. Motivated by this observation, we propose a pixel neighborhood cross-attention mechanism. This localized and dynamic interaction strategy reduces redundant feature matching and improves the memory/FLOPs trade-off for resource-constrained multispectral perception, although~absolute latency still depends on hardware support for deformable~sampling.

The main contributions of this paper are summarized as~follows:
\begin{itemize}
	\item We propose PNAFusion, a~progressive pixel-neighborhood deformable cross-attention framework for weakly aligned multispectral object detection. The~framework explicitly addresses the trade-off between cross-modal alignment accuracy and memory/FLOPs scalability in high-resolution feature fusion, rather than claiming a strict latency advantage.
	\item We design Adaptive Deformable Alignment (ADA) and Pixel-Neighborhood Cross-Attention (PNCA) modules. ADA learns content-aware sampling offsets to compensate for local non-linear misalignment, while PNCA restricts cross-modal interaction to a $k \times k$ neighborhood and reduces the dominant attention complexity from $\mathcal{O}((HW)^2)$ to $\mathcal{O}(HWk^2)$.
	\item Extensive experiments on FLIR, M3FD, and~DroneVehicle demonstrate that PNAFusion achieves highly competitive detection accuracy against current multispectral detection frameworks. Compared with global cross-attention, it substantially reduces allocated GPU memory, while its additional latency caused by deformable sampling and iterative refinement is explicitly analyzed.
\end{itemize} 

\par 
The rest of this paper is organized as follows. Section~\ref{sec2} reviews related work on multispectral object detection and attention-based methods. Section~\ref{sec3} describes the proposed PNAFusion framework. Section~\ref{sec4} presents experimental results and analysis. Section~\ref{sec5} concludes the paper and discusses~limitations.

\section{Related~Works}
\label{sec2}

\subsection{Multispectral Object~Detection}

Multispectral object detection improves the robustness of detectors under complex illumination conditions and adverse weather by integrating the rich texture details of visible-spectrum (RGB) images with the thermal radiation characteristics of infrared (TIR) images. According to the evolution of feature interaction paradigms, existing studies can be broadly categorized into three groups: CNN-based methods~\cite{ref12}, Transformer-based methods, and~Mamba-based methods~\cite{ref13,ref14,ref15}.

Early dual-stream fusion architectures were primarily based on convolutional neural networks, which required carefully designed feature fusion modules to integrate features from different modalities. Zhang~et~al.~\cite{ref16} proposed Guided Attentive Feature Fusion (GAFF), which adaptively regulates the contribution of different modalities through weighted feature aggregation. Zhou~et~al.~\cite{ref17} introduced an illumination-aware module that dynamically adjusts the fusion strategy according to the illumination distribution. However, due to the limited local receptive field of convolutional operators, such methods are inadequate for modeling long-range interactions. Moreover, conventional CNN-based frameworks usually assume that the input images are pre-aligned. When such architectures encounter the ``weak misalignment'' problem caused by physical offsets between sensors, issues such as feature ghosting and localization errors often~arise.

With the introduction of self-attention and cross-attention mechanisms, this class of models has demonstrated superior capability in modeling global contextual relationships, and~Vision Transformers (ViTs) have been widely applied to multispectral fusion tasks in recent years. Qing~et~al.~\cite{ref5} proposed the Cross-modality Fusion Transformer (CFT), which exploits a global attention mechanism to capture long-range complementary information between RGB and TIR modalities. ICAFusion~\cite{ref7} introduced an iterative cross-attention module that progressively refines fused features through a multi-stage interaction strategy. Although~Transformer architectures perform well in modeling complex semantic interactions across modalities, their computational complexity increases quadratically with image resolution, i.e.,~$O(H^2W^2)$. This not only imposes substantial memory overhead, but~also limits the feasibility of deploying such models on low-power platforms such as embedded devices. In~addition, global interaction may introduce background noise irrelevant to the target features, thereby impairing object~recognition.

Beyond improvements in the spatial domain, researchers have also begun to explore the value of frequency-domain information in multispectral fusion. Zuo~et~al.~\cite{ref19} proposed the SFFR method, which employs a Kolmogorov--Arnold Network (KAN) to reconstruct features jointly in the spatial and frequency domains. Through selective frequency component exchange, this method is able to capture subtle cross-modal consistency that is difficult to discover using conventional convolution or attention mechanisms. This spatial--frequency collaborative design offers a new direction for improving model robustness in complex aerial~scenarios.

To address the computational bottleneck introduced by Transformers, recent research has shifted toward state space models (SSMs) with linear computational complexity. Shen~et~al.~\cite{ref20} proposed MS2Fusion, which achieves an effective balance between shared semantic extraction and complementary feature modeling through a dual-path parameter interaction mechanism. Dong~et~al.~\cite{dong_fusion-mamba_2024} proposed Fusion-Mamba, which performs cross-modal interaction in the hidden state space through a gating mechanism, effectively suppressing interference from false target information. Although~Mamba-based methods significantly improve long-sequence modeling efficiency, their one-dimensional scanning characteristic still faces challenges in handling precise two-dimensional spatial alignment, such as pixel-level deformable correction.
In this paper, we propose PNAFusion, which leverages the precise interaction capabilities of Transformers while simultaneously optimizing both computational efficiency and fusion accuracy through a novel neighborhood deformable~mechanism.

\subsection{Attention-Based~Methods}

As the core component of the Transformer architecture, the~self-attention mechanism overcomes the limitations of the local receptive field in traditional convolutional neural networks. It enables global long-range interaction by computing correlation scores among all elements in a sequence. Specifically, this mechanism maps the input features into a query matrix (Query), a~key matrix (Key), and~a value matrix (Value), and~determines the attention weights by computing the dot product between Query and Key. In~this way, the~model can dynamically focus on the most important regions in an image according to the content. Owing to this powerful capability for capturing long-range dependencies, Transformers can better understand semantic relationships in complex scenes for object detection~tasks.

To apply this capability to multimodal scenarios, the~cross-attention mechanism has been widely used for the dynamic selection and integration of cross-modal information. In~general, cross-attention treats one modality (e.g., RGB) as the Query and computes its relevance to the Key and Value derived from another modality (e.g., TIR). Attention mechanisms, including self-attention and cross-attention, have become core techniques in computer vision for capturing long-range dependencies~\cite{ref22,ref23}.

Although global attention mechanisms can theoretically establish semantic associations over the entire image, their substantial computational and memory burden poses severe challenges in practical applications. To~address these issues, many improved variants have been proposed. For~example, Swin Transformer~\cite{ref24} restricts computation to local windows through a shifted-window mechanism, achieving a trade-off between computational efficiency and receptive field. Neighborhood Attention (NAT)~\cite{ref25} further constrains the interaction range to pixel-level neighborhoods, which not only preserves translation equivariance but also significantly reduces computational~cost.

When dealing with the ``weak misalignment'' problem in multispectral images, traditional fixed attention windows are often inadequate for precise feature alignment. Inspired by deformable convolution (DCN), Zhu~et~al.~\cite{ref26} proposed deformable attention, which dynamically predicts sampling offsets to guide the network toward more discriminative local regions. This strategy provides a new perspective for addressing the ``weak misalignment'' problem caused by sensor bias. However, how to achieve content-aware dynamic deformable alignment while preserving localized neighborhood interaction remains a key challenge in efficient multispectral fusion. Motivated by this line of development, this paper proposes a pixel-neighborhood cross-attention module with adaptive deformable alignment capability, aiming to achieve high-precision cross-modal feature correction and fusion within a neighborhood-constrained search space. In~addition to spatial attention in image-level fusion, progressive refinement has also been explored in related dense prediction tasks. Dong~et~al.~\cite{dong2024learning} proposed Learning Temporal Distribution and Spatial Correlation (LTS) for universal moving object segmentation, where a Defect Iterative Distribution Learning (DIDL) network learns temporal pixel distributions and a Stochastic Bayesian Refinement (SBR) network further models spatial correlation. Although~LTS focuses on video moving object segmentation rather than visible--thermal object detection, it shares a relevant design principle with our work: both methods avoid relying on a one-shot correspondence estimation and instead adopt progressive refinement to improve spatial consistency. The~key difference is that LTS refines temporal--spatial foreground masks in videos, whereas PNAFusion refines cross-modal feature alignment between visible and thermal images through neighborhood-constrained deformable sampling and~cross-attention. 

\section{Method}
\label{sec3}

\subsection{Overview}
\begin{figure}[H]
	\centering
	\includegraphics[width=\textwidth]{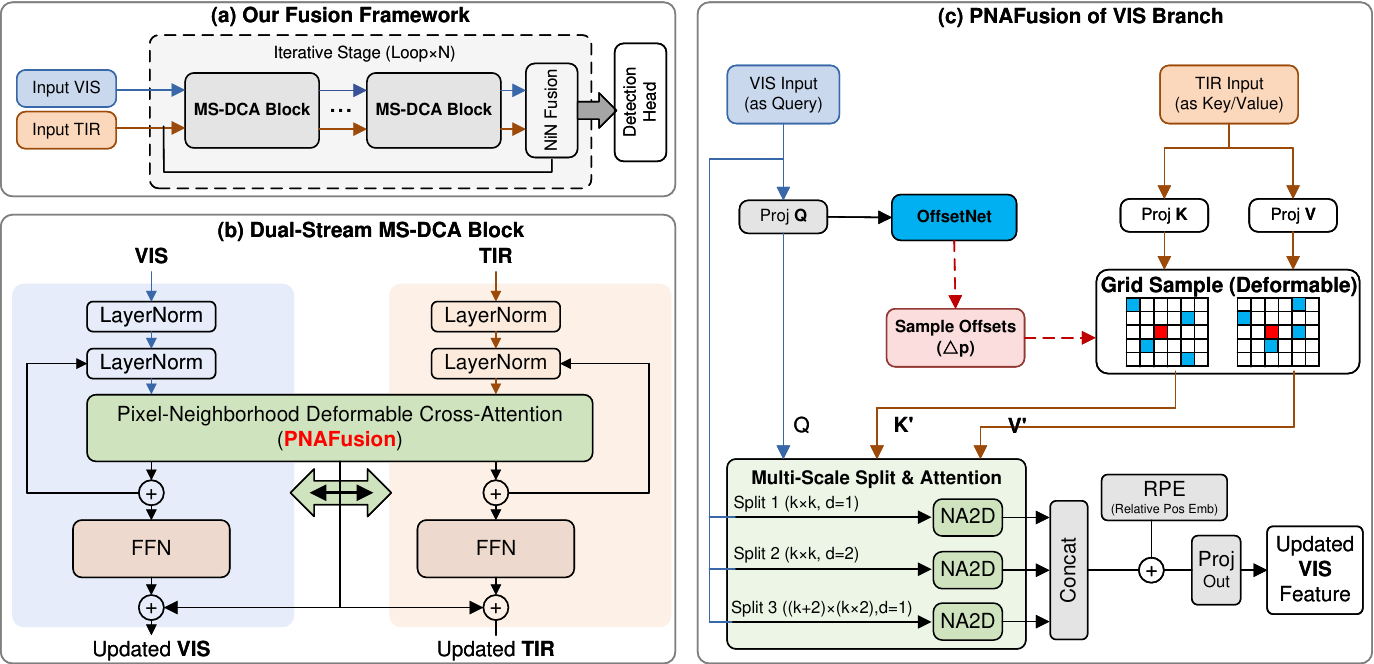}
	\caption{{Overall} 
		{architecture} 
		of the proposed framework. (\textbf{a}) It consists of three components: a dual-stream feature extraction module, dual-stream MS-DCA blocks and a detection head; \mbox{(\textbf{b}) dual-stream} MS-DCA blocks for feature interaction; (\textbf{c}) a detection head for regression of object location, class label and confidence~score.}
	\label{fig:framework}
\end{figure}

This paper proposes a dual-stream multispectral object detection framework that achieves efficient feature fusion through local interaction and adaptive alignment within the network. The~overall architecture is illustrated in Figure~\ref{fig:framework}. It consists of three main stages: a dual-stream feature extraction stage, an~Iterative Pixel-Neighborhood Deformable Cross-Attention fusion stage and a detection~Head.

As shown in Figure~\ref{fig:framework}a, the~proposed architecture first employs parallel backbone networks to extract multi-scale features from the RGB (VIS) and infrared (TIR) modalities. It adopts a hierarchically integrated multi-scale fusion strategy embedded within the backbone, where the fusion modules are deployed at feature levels with downsampling rates of $\{8,\,16,\,32\}$, corresponding to the P3, P4, and~P5 stages.
Subsequently, the~multi-scale features are fed into a fusion stage equipped with an iterative refining mechanism (Loop $N$), and~are finally compressed by NiNFusion ($1\times1$ convolution) to produce object representations for the detection~head.

As shown in Figure~\ref{fig:framework}b, within~each feature level of P3, P4, and~P5, the~Dual-Stream MS-DCA block relies on the alternating stacking of dual-stream multispectral deformable Transformer blocks to achieve deep semantic interaction across modalities.
In Figure~\ref{fig:framework}c, the~core mechanism tightly integrates adaptive content-aware pre-alignment with local-window~cross-attention. 

\subsection{Difference from Related Attention and Deformation Operators} 

PNAFusion is not a simple cascade of Neighborhood Attention, deformable attention, and~deformable convolution. Standard Neighborhood Attention Transformer (NAT) computes self-attention within a fixed local neighborhood of a single modality. It preserves local translation equivariance and reduces attention cost, but~it does not explicitly perform cross-modal Query--Key--Value matching and does not predict spatial offsets for modality-specific displacement. Deformable convolution and deformable attention introduce learnable sampling offsets, but~they are usually designed for single-modality feature adaptation or multi-scale sparse sampling and do not explicitly constrain cross-modal correspondence within a semantically related pixel neighborhood. In~contrast, PNAFusion tightly couples ADA and PNCA: ADA predicts content-aware offsets for the key-value modality, and~PNCA uses the aligned key-value features to perform local cross-modal attention. The~deformable search space is restricted to a $k \times k$ neighborhood around each query pixel, which suppresses irrelevant global background matching while retaining the ability to correct local non-linear misalignment between visible and thermal~features. 

\subsection{Pixel-Neighborhood Cross-Attention (PNCA)} 
\label{sec3.3} 
Although conventional global cross-attention can capture long-range dependencies, its computational cost for high-resolution multispectral feature maps grows quadratically as $\mathcal{O}((HW)^2)$. Moreover, in~weakly aligned multispectral fusion, useful complementary information is often concentrated around the target and its adjacent spatial neighborhood, whereas global search may introduce irrelevant background responses. To~reduce redundant matching and suppress background noise, we propose Pixel-Neighborhood Cross-Attention (PNCA), which restricts cross-modal interaction to a local window centered at each query position. Let the query-modality feature and the key-value-modality feature be denoted {as} 
\begin{equation} 
X_q \in \mathbb{R}^{H \times W \times C}, \qquad X_{kv} \in \mathbb{R}^{H \times W \times C}, 
\end{equation} 
where $H$, $W$, and~$C$ denote the feature height, width, and~channel dimension, respectively. After~linear projection, the~features are divided into $M$ attention heads, and~the channel dimension of each head is $d=C/M$. For~a query position $i=(x,y)$, the~projected query vector is
\begin{equation}
Q_i \in \mathbb{R}^{1 \times d}. 
\end{equation} 

{After} 
the ADA module described in Section~\ref{sec3.3}, the~aligned key and value vectors at a neighboring position $j$ are denoted as
\begin{equation} 
K'_j \in \mathbb{R}^{1 \times d}, \qquad V'_j \in \mathbb{R}^{1 \times d}. 
\end{equation} 

For each query position $i=(x,y)$, PNCA defines a $k \times k$ neighborhood as
\begin{equation} 
\mathcal{N}(i)= \left\{ j=(x',y'):\ |x-x'|\leq \left\lfloor \frac{k}{2} \right\rfloor,\ |y-y'|\leq \left\lfloor \frac{k}{2} \right\rfloor \right\}. 
\end{equation} 

The attention output at position $i$ is computed as
\begin{equation} 
\operatorname{PNCA}(Q,K',V')_i = \sum_{j\in \mathcal{N}(i)} 
\operatorname{Softmax}_{j\in \mathcal{N}(i)} \left( \frac{Q_i(K'_j)^{\top}+B_{i,j}}{\sqrt{d}} \right)V'_j,
\end{equation} 
where $B_{i,j}$ is the relative position bias between the query position $i$ and the neighboring key position $j$. This bias compensates for the positional information weakened by localized interaction. Compared with global cross-attention, PNCA reduces the dominant complexity from
\begin{equation} 
\Omega_{\mathrm{Global}} = 2(HW)^2C 
\end{equation} to
\begin{equation} 
\Omega_{\mathrm{PNCA}} = 2HWk^2C. 
\end{equation} 

Since $k \ll \min(H,W)$ in practical feature maps, the~proposed local cross-modal interaction substantially reduces memory consumption while preserving the ability to model local non-linear~relationships.

\subsection{Adaptive Deformable Alignment (ADA)} 
Visible and thermal images are often weakly misaligned due to sensor displacement, calibration residuals, and~modality-specific imaging geometry. Directly computing cross-attention on such features may aggregate semantically inconsistent regions and produce feature ghosting. To~address this issue, we introduce Adaptive Deformable Alignment (ADA) before PNCA. ADA predicts local content-aware offsets and resamples the key-value modality so that semantically corresponding features are better aligned before cross-modal attention. Given $X_q \in \mathbb{R}^{H \times W \times C}$ and $X_{kv}\in \mathbb{R}^{H \times W \times C}$, an~offset generation network predicts pixel-wise offsets:
\begin{equation} 
\Delta p = \operatorname{OffsetNet}\left([X_q,X_{kv}]\right), 
\end{equation} 
where $[\cdot,\cdot]$ denotes channel-wise concatenation. For~a $k \times k$ neighborhood, the~offset tensor is defined as
\begin{equation} 
\Delta p \in \mathbb{R}^{H \times W \times 2k^2}, 
\end{equation} 
where the factor $2$ corresponds to horizontal and vertical coordinate displacements for each sampling point in the local neighborhood. The~aligned key-value feature $X'_{kv}$ is obtained by deformable bilinear sampling:
\begin{equation} 
X'_{kv}(p)= \sum_{q}G(q,p+\Delta p)\cdot X_{kv}(q), 
\end{equation} 
where $p$ is a sampling position, $q$ enumerates discrete positions on the original feature map, and~$G(\cdot,\cdot)$ denotes the bilinear interpolation kernel. The~aligned feature $X'_{kv}$ is then projected into $K'$ and $V'$ and used by PNCA:
\begin{equation} 
F_{\mathrm{aligned}}= \operatorname{PNDCA}\left(Q,K(p+\Delta p),V(p+\Delta p)\right). 
\end{equation} 

This design differs from unconstrained deformable attention because the offsets are used inside a neighborhood-constrained cross-modal attention window. Therefore, ADA does not perform arbitrary global sampling. Instead, it corrects local non-linear displacement while keeping feature aggregation focused on semantically related \mbox{neighboring~regions.} 

\subsection{Progressive Refinement with Iterative~Feedback} 
A single cross-modal interaction may be insufficient to fully capture complementary semantic cues or eliminate weak spatial misalignment. Therefore, PNAFusion introduces an iterative feedback mechanism that progressively refines visible and thermal features. Let the initial multispectral features extracted by the backbone be denoted as $F^{(0)}_{\mathrm{vis}}$ and $F^{(0)}_{\mathrm{tir}}$. At~the $n$-th iteration, the~two feature streams are updated by a dual-stream PNDCA block:
\begin{equation} 
\left( F^{(n+1)}_{\mathrm{vis}}, F^{(n+1)}_{\mathrm{tir}} \right) = \Phi_{\mathrm{PNDCA}} \left( F^{(n)}_{\mathrm{vis}}, F^{(n)}_{\mathrm{tir}} \right), 
\end{equation} 
where $\Phi_{\mathrm{PNDCA}}(\cdot)$ denotes the cross-modal interaction block that jointly incorporates ADA and PNCA. Through recurrent feedback, the~offset estimation and local cross-modal attention are repeatedly updated, allowing the model to progressively reduce residual alignment errors. This progressive strategy is conceptually related to iterative refinement designs in dense prediction. For~example, LTS~\cite{dong2024learning} progressively learns temporal distribution and spatial correlation for universal moving object segmentation. However, the~task and mechanism are different: LTS refines video foreground masks through temporal distribution learning and spatial Bayesian refinement, whereas PNAFusion refines visible--thermal feature correspondence through neighborhood-constrained deformable cross-attention. After~$N$ recurrent iterations, the~model obtains two semantically synchronized feature streams, $F^{(N)}_{\mathrm{vis}}$ and $F^{(N)}_{\mathrm{tir}}$. We then use a lightweight Network-in-Network (NiN) fusion module for final aggregation. Specifically, the~two refined feature streams are concatenated along the channel dimension and compressed by a $1\times1$ convolution:
\begin{equation} 
F_{\mathrm{fused}} = \operatorname{Conv}_{1\times1} \left( \left[ F^{(N)}_{\mathrm{vis}}, F^{(N)}_{\mathrm{tir}} \right] \right). 
\end{equation} 

Here, $[\cdot,\cdot]$ denotes channel-wise concatenation. The~$1\times1$ convolution is selected because the preceding PNDCA blocks have already performed spatial alignment and local semantic interaction. At~this stage, the~main purpose is channel mixing and dimensional compression rather than another expensive spatial interaction. Compared with adding another attention-based fusion module, NiN fusion is more lightweight and preserves the local structural correspondence established by~PNCA. 

\subsection{Loss~Function}
To optimize alignment performance and detection accuracy, we introduce a multi-task loss function. The~total loss $L_{\text{total}}$ is defined as a weighted combination of the bounding box regression loss $L_{\text{box}}$, objectness loss $L_{\text{obj}}$, and~classification loss $L_{\text{cls}}$:
\begin{equation}
L_{\text{total}}=\lambda_1 L_{\text{box}}+\lambda_2 L_{\text{obj}}+\lambda_3 L_{\text{cls}},
\end{equation}
where $\lambda_1$, $\lambda_2$, and~$\lambda_3$ are hyperparameters used to balance different tasks. Through detection supervision, the~offset generation network in ADA learns spatial alignment parameters that maximize detection accuracy. This training scheme ensures that the proposed model can maintain optimal detection performance and alignment precision even in complex low-power real-time application~scenarios.

\subsection{Computational~Complexity} 
To analyze the efficiency of PNAFusion, we compare the dominant computational cost of global cross-attention and the proposed neighborhood-constrained deformable cross-attention. Let a feature map at level $l$ have spatial size $H_l \times W_l$ and channel dimension $C_l$. For~global cross-attention, the~dominant cost comes from full-image Query--Key correlation and attention-weighted Value aggregation:
\begin{equation} 
\Omega^{(l)}_{\mathrm{Global}} \approx 2(H_lW_l)^2C_l. 
\end{equation} 

This quadratic growth becomes costly for high-resolution feature maps, especially at the P3 level. In~contrast, PNCA restricts cross-modal interaction to a $k \times k$ local neighborhood. The~corresponding attention complexity at feature level $l$ is
\begin{equation} 
\Omega^{(l)}_{\mathrm{PNCA}} \approx 2H_lW_lk^2C_l. 
\end{equation} 

ADA additionally introduces an OffsetNet for predicting local sampling offsets. If~the offset generator contains convolutional layers indexed by $s$, with~kernel size $K_s$, input channel dimension $C^{\mathrm{in}}_{l,s}$, and~output channel dimension $C^{\mathrm{out}}_{l,s}$, its computational cost at feature level $l$ can be written as
\begin{equation} 
\Omega^{(l)}_{\mathrm{ADA}} \approx \sum_{s} H_lW_lK_s^2C^{\mathrm{in}}_{l,s}C^{\mathrm{out}}_{l,s}. 
\end{equation} 

Therefore, the~total complexity of PNAFusion with $N$ iterative refinement rounds over feature levels $\mathcal{L}=\{P3,P4,P5\}$ is
\begin{equation} 
\Omega_{\mathrm{Total}} = N\cdot \sum_{l\in\mathcal{L}} \left( \Omega^{(l)}_{\mathrm{PNCA}} + \Omega^{(l)}_{\mathrm{ADA}} \right). 
\end{equation} 

This formulation explicitly includes the repeated cost of the offset generator. Although~OffsetNet is lightweight compared with the main attention operation, its cost is accumulated over feature levels and recurrent iterations. Thus, the~theoretical analysis is more accurate than treating offset prediction as negligible. Since $k \ll \min(H_l,W_l)$, the~attention term remains linear with respect to the number of pixels, while the practical latency is also affected by the hardware efficiency of deformable bilinear~sampling. 

\section{{Experiments}} 
\label{sec4}

\subsection{Experimental~Setting} 
We evaluate PNAFusion on three public multispectral object detection benchmarks: FLIR, M3FD, and~DroneVehicle. For~FLIR, we follow the train--test split used in ICAFusion to ensure comparability. For~M3FD and DroneVehicle, we follow the official or commonly adopted public splits used by the compared methods. No additional manual day/night relabeling or custom split is introduced. All input images are resized to $640\times640$. The~implementation is based on PyTorch 2.0. Unless~otherwise specified, experiments are conducted on a server equipped with an NVIDIA GeForce RTX 4090D GPU with 24 GB memory, an~AMD EPYC 9754 128-Core Processor, and~60 GB system memory. The~model is trained end-to-end for 100 epochs using the Adam optimizer. The~initial learning rate is set to $1.0\times10^{-3}$ and decayed by cosine annealing. The~weight decay is $5.0\times10^{-4}$. The~batch size is set to 16. The~random seed is fixed to 42 for reproducibility. The~adopted data augmentations include Mosaic, MixUp, and~random horizontal flipping. The~detection loss follows the YOLO-style multi-task objective:
\begin{equation} 
\mathcal{L}_{\mathrm{total}} = \lambda_1\mathcal{L}_{\mathrm{box}} + \lambda_2\mathcal{L}_{\mathrm{obj}} + \lambda_3\mathcal{L}_{\mathrm{cls}}, 
\end{equation} 
where $\mathcal{L}_{\mathrm{box}}$, $\mathcal{L}_{\mathrm{obj}}$, and~$\mathcal{L}_{\mathrm{cls}}$ denote the bounding-box regression loss, objectness loss, and~classification loss, respectively. The~loss weights are set to $\lambda_1=0.05$, $\lambda_2=1.0$, and~$\lambda_3=0.5$. Performance is measured under the COCO evaluation protocol using mAP@0.5, mAP@0.75, and~mAP@0.5:0.95, together with category-level AP when~available. 

\subsection{Comparison with State-of-the-Art~Methods}

In this section, we comprehensively compare the proposed PNAFusion with mainstream methods in multispectral object detection to validate its superiority in both detection accuracy and computational efficiency. The~compared methods include classical CNN-based fusion models, representative Transformer-based approaches, as~well as our baseline model ICAFusion. To~make the comparison more transparent, Tables~\ref{tab:flir_comparison}--\ref{tab:dronevehicle_comparison} explicitly report the detector/backbone setting and result source in a separate column. Entries marked as ``Published'' are directly taken from the corresponding papers, whereas entries marked as ``Ours'' are reproduced or implemented by the authors under our experimental~pipeline.

\begin{table}[H]
\footnotesize
	\centering
	\caption{Comparison with state-of-the-art methods on the FLIR dataset.}
	\label{tab:flir_comparison}
	\begin{tabularx}{\textwidth}{l l c c c c c}
		\toprule
		\textbf{Method} & \textbf{Detector/Source} & \textbf{mAP@0.5} & \textbf{mAP} & \textbf{Person} & \textbf{Car} & \textbf{Bicycle} \\
		\midrule
		GAFF~\cite{ref16} & Published & 72.9  & -- & -- & -- & -- \\
		CFT~\cite{ref5} & Published & 78.7  & -- & -- & -- & -- \\
		CSAA~\cite{csaa_cao2023multimodal} & Published & 79.2  & -- & -- & -- & -- \\
		ICAFusion~\cite{ref7} & Published & 79.2  & 41.4 & 81.6 & 89.0 & 66.9 \\
		MMFN~\cite{mmfn_yang2024multidimensional} & Published & 80.8 &  41.7 & 85.7 & 91.2 & 65.5 \\
		CPCF~\cite{CPCF_hu2024rethinking} & Published & 82.1  & -- & -- & -- & -- \\
		GM-DETR~\cite{xiao2024gm} & Published & 83.9  & -- & -- & -- & -- \\
		Fusion-Mamba~\cite{dong_fusion-mamba_2024} & Published & 84.3  & -- & -- & -- & -- \\
		TFDet~\cite{zhang2024tfdet} & Published & 86.6 & 46.6 & -- & -- & -- \\
		DAMSDet~\cite{guo2024damsdet} & Published & 86.6 & 49.3 & -- & -- & -- \\
		PNAFusion & YOLOv5, Ours & 84.2 & 43.8 & 87.2 & 91.1 & 74.3 \\
		PNAFusion & Co-DETR, Ours & 86.8 & 50.3 & 90.4 & 93.4 & 76.5 \\
		\bottomrule
	\end{tabularx}
\end{table}
\unskip

\begin{table}[H]
	\centering
	\footnotesize  
	\caption{Comparison with state-of-the-art methods on the M3FD~dataset.}
	\label{tab:m3fd_comparison}
	\begin{tabular}{l l c c c c c c c c}
		\toprule
		\textbf{Method} & \textbf{Detector/Source} & \textbf{mAP@0.5} & \textbf{mAP} & \textbf{People} & \textbf{Car} & \textbf{Bus} & \textbf{Lamp} & \textbf{Motor} & \textbf{Truck} \\
		\midrule
		ICAFusion~\cite{ref7} & Published & 67.8 & 41.9 & -- & -- & -- & -- & -- & -- \\
		CFT~\cite{ref5} & Published & 68.2 & 42.5 & -- & -- & -- & -- & -- & -- \\
		DAMSDet~\cite{guo2024damsdet} & Published & 80.2 & 52.9 & -- & -- & -- & -- & -- & -- \\
		SuperFusion~\cite{tang2022superfusion} & Published & 83.5 & 57.0 & 83.7 & 91.0 & 93.2 & 70.0 & 77.4 & 85.8 \\
		MMFN~\cite{mmfn_yang2024multidimensional} & Published & 86.2 & -- & 83.0 & 93.2 & 92.1 & 87.6 & 73.7 & 87.4 \\
		Fusion-Mamba~\cite{dong_fusion-mamba_2024} & Published & 88.0 & 61.9 & 84.3 & 92.9 & 94.2 & 87.5 & 80.5 & 88.8 \\
		PNAFusion & YOLOv5, Ours & 90.5 & 62.4 & 87.8 & 94.1 & 94.0 & 93.5 & 82.6 & 91.2 \\
		PNAFusion & Co-DETR, Ours & \textbf{90.8} & \textbf{62.9} & 90.2 & 94.7 & 94.5 & 87.8 & 85.6 & 91.9 \\
		\bottomrule
	\end{tabular}
\end{table}

\begin{table}[H]
	\footnotesize
	\caption{Comparison with state-of-the-art methods on the DroneVehicle~dataset.}
	\label{tab:dronevehicle_comparison}
	\begin{tabularx}{\textwidth}{llcccccc}
		\toprule
		\textbf{Method} & \textbf{{Detector/Source}} & \textbf{mAP@0.5} & \textbf{Car} & \textbf{Truck} & \textbf{Bus} & \textbf{Van} & \textbf{Freight~Car} \\
		\midrule
		DCCINet~\cite{bao_dual-dynamic_2025} & Published & 78.4 & 91.0 & 78.9 & 90.7 & 65.5 & 66.1 \\
		ICAFusion~\cite{ref7} & Published & 78.6 & 96.7 & 79.0 & 95.8 & 61.8 & 60.0 \\
		CCLDet~\cite{shang_ccldet_2025} & Published & 79.4 & 97.7 & 75.4 & 95.7 & 59.5 & 68.8 \\
		DMM~\cite{zhou_dmm_2024} & Published & 79.4 & 90.4 & 79.8 & 89.9 & 68.6 & 68.2 \\
		MGMF~\cite{liu_como_2026} & Published & 80.3 & 91.4 & 78.5 & 91.1 & 69.4 & 70.1 \\
		RGFNet~\cite{zhao_rgfnet_2025} & Published & 81.4 & 98.4 & 81.1 & 95.8 & 63.0 & 68.7 \\
		PNAFusion & YOLOv5, Ours & \textbf{85.5} & 98.1 & 85.0 & 97.2 & 71.7 & 75.8 \\
		\bottomrule
	\end{tabularx}
\end{table}
\unskip

\subsubsection{Results on the FLIR~Dataset}

Table~\ref{tab:flir_comparison} presents the quantitative comparison on the FLIR dataset. Under~the YOLOv5 detector, PNAFusion achieves 84.2 mAP@0.5, which is substantially higher than the YOLOv5-based ICAFusion baseline and remains comparable to Fusion-Mamba. It should be noted that direct comparison across different detector families is not strictly controlled, because~DETR-based frameworks such as TFDet and DAMSDet usually benefit from stronger global context modeling and larger detector capacity. Therefore, we report both PNAFusion (YOLOv5) and PNAFusion (Co-DETR) to clarify two aspects: the YOLOv5 result reflects fair comparison with YOLO-style multispectral fusion baselines, while the Co-DETR result demonstrates that the proposed fusion module can also be transferred to a stronger Transformer-based detector. In~terms of high-precision localization, PNAFusion (YOLOv5) achieves 39.4 mAP@0.75, outperforming ICAFusion. This improvement is important because mAP@0.75 is more sensitive to spatial alignment and bounding-box tightness than mAP@0.5. The~result indicates that the proposed ADA and PNCA modules help reduce feature ghosting and localization bias under weak cross-modal misalignment. When integrated into Co-DETR, PNAFusion further reaches 86.8 mAP@0.5 and 50.3 mAP@0.5:0.95, showing that the proposed fusion design has cross-framework~scalability. 

\subsubsection{Results on the M3FD~Dataset}

Table~\ref{tab:m3fd_comparison} reports the detection performance on the M3FD dataset. PNAFusion achieves 90.5 mAP@0.5 under the YOLOv5 detector and 90.8 mAP@0.5 when transferred to Co-DETR. These results indicate that the proposed neighborhood-constrained deformable interaction is effective not only for weakly aligned traffic scenes but also for relatively well-aligned multispectral image pairs. At~the category level, PNAFusion obtains competitive or superior performance on several classes, especially categories with distinctive thermal responses or small object scales. This suggests that progressive local cross-modal refinement can enhance both semantic discrimination and localization~robustness.

\subsubsection{Results on the DroneVehicle~Dataset}

Table~\ref{tab:dronevehicle_comparison} shows the comparison on the DroneVehicle dataset. This benchmark contains high-resolution aerial visible--infrared image pairs with small object scales and complex backgrounds, making it challenging for cross-modal fusion. PNAFusion achieves \mbox{85.5 mAP@0.5,} outperforming the compared multispectral detection methods. The~improvement can be attributed to the local deformable cross-attention mechanism: PNCA suppresses irrelevant background matching by restricting interaction to pixel neighborhoods, while ADA compensates for local spatial offsets between modalities. These properties are particularly beneficial for aerial scenes, where small targets are easily affected by background clutter and weak registration~errors.

\subsection{Ablation~Studies}

To quantitatively analyze the effects of the core components and hyperparameters of PNAFusion on detection performance, we conduct a series of ablation studies on the FLIR dataset. We focus on the contributions of neighborhood window size, iteration rounds, and~adaptive offsets. Except~for the tested variable, all other experimental settings are kept~unchanged.

\subsubsection{Different Number of~Blocks}

In the PNAFusion architecture, the~number of stacked Multi-Transformer Blocks, denoted by $L$, represents the depth of cross-modal feature interaction. Increasing $L$ can theoretically enhance the model's semantic modeling capability in complex scenes, but~it also leads to a substantial increase in computational cost. To~investigate the intrinsic relationship between model depth and detection performance, we conduct comparative experiments with $L\in\{2,3,4,5,6\}$. The~results are shown in Table~\ref{tab:num_blocks}.

\begin{table}[H]
	\footnotesize
	\centering
	\caption{Ablation study on different numbers of Multi-Transformer Blocks $L$ (kernel size = 3) on the FLIR~dataset.}
	\label{tab:num_blocks}
	\begin{tabular}{c c c c c c c c c c c c c c}
		\toprule
		\multirow{2}{*}{\textbf{Num Blocks}} 
		& \multicolumn{3}{c}{\textbf{mAP@0.5}} 
		& \multicolumn{3}{c}{\textbf{mAP@0.75}} 
		& \multicolumn{3}{c}{\textbf{mAP}} 
		& \multirow{2}{*}{\textbf{Params (M)}} \\
		\cmidrule(lr){2-4} \cmidrule(lr){5-7} \cmidrule(lr){8-10}
		& \textbf{Person} & \textbf{Car} & \textbf{Bicycle} 
		& \textbf{Person} & \textbf{Car} & \textbf{Bicycle} 
		& \textbf{Person} & \textbf{Car} & \textbf{Bicycle} & \\
		\midrule
		2 & 85.0 & 89.9 & 71.0 & 25.7 & 61.5 & 8.9 & 37.8 & 56.9 & 25.9 & 130.9 \\
		3 & 83.0 & 89.6 & 72.6 & 26.5 & 61.5 & 9.3 & 36.8 & 57.0 & 25.3 & 158.6 \\
		4 & 83.9 & 90.4 & 71.4 & 30.2 & 61.6 & 7.8 & 38.7 & 56.6 & 23.6 & 186.3 \\
		5 & 81.3 & 88.5 & 72.6 & 15.0 & 53.4 & 10.3 & 31.5 & 50.5 & 24.3 & 214.1 \\
		6 & 82.9 & 88.6 & 71.8 & 27.7 & 56.8 & 9.8 & 37.2 & 52.1 & 26.8 & 241.7 \\
		\bottomrule
	\end{tabular}
\end{table}

Sufficient shallow feature interaction. As~shown in Table~\ref{tab:num_blocks}, the~model achieves the best overall performance when $L=2$. This indicates that, after~introducing the iterative feedback mechanism, PNAFusion can repeatedly refine features by recurrently invoking the same feature interaction module, thereby achieving highly accurate alignment even with relatively limited physical depth. Compared with simply increasing the stacking depth, this iterative mechanism is more effective for handling the weak misalignment~problem.

Feature saturation and degradation. As~$L$ increases from 3 to 6, the~detection performance does not improve linearly, but~instead exhibits a fluctuating downward trend. For~example, when $L=4$, the~overall mAP$_{50}$ decreases by 0.1 compared with the $L=2$ setting, and~the overall mAP drops from 40.2 to 39.6. This phenomenon can be attributed to feature oversmoothing and the increased difficulty of optimization. On~the one hand, excessive cross-modal interaction tends to average feature representations, thereby reducing feature contrast. On~the other hand, under~the limited scale of the FLIR dataset, the~model size increases from 130.9M to 241.7M parameters, which significantly raises the risk of overfitting and aggravates gradient instability during deep network~training.

Considering both the need for sufficient feature fusion depth and the computational constraints of practical hardware platforms, we finally adopt $L=2$ as the default configuration of~PNAFusion.

\subsubsection{Different Sizes of Pixel~Neighborhood}

The PNDCA mechanism reduces computational cost and resource redundancy by dynamically adjusting the size of the attention window. The~neighborhood window size $k$ is a key hyperparameter of this module, as~it directly determines the model's ability to handle cross-modal spatial misalignment. To~obtain the optimal model performance, it is necessary to identify the most suitable local search radius. Therefore, we conduct comparative experiments with three neighborhood sizes, i.e.,~$k\in\{3,5,7\}$. The~results are shown in Table~\ref{tab:kernel_size}.

\begin{table}[H]
	\footnotesize
	\caption{{Ablation} 
		study on different pixel neighborhood sizes $k$ with {$\texttt{num\_blocks}=2$.} 
	}
	\label{tab:kernel_size}
	\begin{tabularx}{\textwidth}{ccccccccc}
		\toprule
		\multirow{2.5}{*}{\textbf{Kernel Size \emph{k}}} 
		& \multicolumn{4}{c}{\textbf{mAP\textsubscript{50}}} 
		& \multicolumn{4}{c}{\textbf{mAP\textsubscript{75}}} \\
		\cmidrule{2-9}
		& \textbf{Person} & \textbf{Car} & \textbf{Bicycle} & \textbf{All} & \textbf{Person} & \textbf{Car} & \textbf{Bicycle} & \textbf{All} \\
		\midrule
		3 & 85.0 & 89.9 & 71.0 & 82.0 & 25.7 & 61.5 & 8.9 & 32.0 \\
		5 & 84.8 & 89.8 & 75.3 & 83.3 & 29.9 & 60.7 & 8.9 & 33.1 \\
		7 & 84.0 & 89.5 & 75.2 & 82.9 & 28.4 & 58.7 & 10.8 & 32.6 \\
		\bottomrule
	\end{tabularx}
\end{table}

From the experimental results, it can be observed that the overall detection accuracy first increases and then declines as the window size $k$ becomes larger. Compared with the smallest $3\times3$ window, the~$5\times5$ neighborhood improves mAP$_{50}$ from 82.0 to 83.3, and~mAP$_{75}$, which reflects stricter localization accuracy, increases more substantially from 32.0 to 33.1. However, when the window is further enlarged to $7\times7$, the~performance drops. These results indicate that the model achieves the best performance with the $5\times5$ neighborhood, which provides sufficient spatial support for the subsequent adaptive deformable alignment~process.

However, when the window becomes excessively large, the~enlarged sampling neighborhood introduces redundant background features and noise interference, which weakens the model's focus on the target core region and increases the difficulty of optimization. In~addition, a~larger neighborhood size leads to higher computational cost, thereby reducing~efficiency.

The above results show that a window size of $5\times5$ provides the best trade-off between spatial realignment capability and feature purity. Therefore, in~the subsequent ablation studies, we adopt $k=5$ as the default setting of~PNAFusion.

\subsubsection{Different Iteration~Numbers}

One of the core characteristics of PNAFusion is the iterative feedback mechanism, which progressively alleviates the weak misalignment problem between modalities through recurrent refinement and thereby enhances feature representation. To~investigate the optimal number of iterations, we further evaluate the effect of $N\in\{1,2,3,4\}$ on model performance under a fixed window size of $k=5$. The~detailed results are presented \mbox{in Table~\ref{tab:iter_num}.}

The results clearly demonstrate the effect of iterative refinement on multispectral fusion. As~$N$ increases from 1 to 3, the~detection accuracy improves steadily. When $N=3$, compared with single-pass interaction ($N=1$), the~overall mAP$_{75}$ increases from 31.9 to 39.4, indicating a substantial improvement in high-precision detection. This suggests that through multiple iterations, the~model can exploit the feedback information generated in the previous fusion round to recursively correct complex nonlinear spatial misalignment, thereby achieving deeper semantic alignment between visible-spectrum texture features and infrared thermal radiation~features.

\begin{table}[H]
	\footnotesize
	\centering
	\caption{Ablation study on different iteration numbers $N$ with $\texttt{num\_blocks}=2$ and $\texttt{kernel\_size}=5$.}
	\label{tab:iter_num}
	\begin{tabular}{c c c c c c c c c c c c c c}
		\toprule
		\multirow{2}{*}{\textbf{Iter}} 
		& \multicolumn{3}{c}{\textbf{mAP@0.5}} 
		& \multicolumn{3}{c}{\textbf{mAP@0.75}} 
		& \multicolumn{3}{c}{\textbf{mAP}} 
		& \multirow{2}{*}{\textbf{Params (M)}} \\
		\cmidrule(lr){2-4} \cmidrule(lr){5-7} \cmidrule(lr){8-10}
		& \textbf{Person} & \textbf{Car} & \textbf{Bicycle} 
		& \textbf{Person} & \textbf{Car} & \textbf{Bicycle} 
		& \textbf{Person} & \textbf{Car} & \textbf{Bicycle} & \\
		\midrule
		1 & 85.0 & 90.8 & 68.7 & 24.1 & 61.4 & 10.1 & 37.5 & 57.1 & 24.4 & -- \\
		2 & 86.1 & 90.2 & 71.5 & 31.4 & 63.3 & 12.3 & 40.5 & 57.9 & 27.0 & 152.9 \\
		3 & 87.2 & 91.1 & 74.3 & 38.4 & 65.6 & 14.2 & 43.4 & 59.4 & 28.8 & -- \\
		4 & 86.2 & 90.3 & 68.1 & 33.0 & 62.2 & 8.5 & 41.1 & 56.9 & 22.3 & -- \\
		\bottomrule
	\end{tabular}
\end{table}

However, when the number of iterations is further increased to $N=4$, the~performance drops significantly, with~mAP$_{50}$ decreasing from 84.2 to 81.6. On~the one hand, excessive iterations increase the optimization difficulty of the model and may lead to gradient instability in deep recurrent refinement. On~the other hand, overly repeated feature interaction may cause over-smoothing, in~which modality-specific discriminative cues are diluted, thereby weakening the model's ability to detect small~objects.

\subsubsection{Effectiveness of Proposed~Components}  
To analyze the contribution of the core components in PNAFusion, we conduct incremental ablation experiments under the optimal hyperparameter configuration, i.e.,~$L=2$, $k=5$, and~$N=3$. We adopt a simple baseline consisting of channel concatenation followed by a $1\times1$ convolution, denoted as NiNFusion. We then compare it with ICAFusion, the~proposed PNDCA block, and~the complete PNAFusion equipped with iterative feedback. The~quantitative results are shown in Table~\ref{tab:component_ablation}. The~results show that direct NiN fusion achieves limited performance because simple channel stacking cannot explicitly model non-linear cross-modal correspondence. ICAFusion improves cross-modal interaction through global cross-attention, but~it still lacks explicit local deformable alignment and introduces large memory overhead. After~replacing global interaction with the proposed PNDCA mechanism, the~model obtains a more localized and memory-scalable fusion pattern. Although~the single-step PNDCA block alone does not fully exploit its alignment potential, adding iterative feedback significantly improves performance. The~complete PNAFusion reaches 84.2 mAP@0.5 and 39.4 mAP@0.75, with~a particularly clear gain on the high-precision localization metric. We do not report ADA and PNCA as two completely independent plug-in modules because they are structurally coupled in the proposed design. ADA predicts local sampling offsets for the key-value modality, while PNCA consumes the aligned key-value features to compute cross-modal attention within the same neighborhood. Removing PNCA would leave ADA without an attention-based semantic aggregation mechanism, whereas removing ADA would reduce PNCA to fixed-window local cross-attention that cannot correct spatial offsets. Therefore, PNDCA is evaluated as a complete functional unit. To~further verify the effect of learned offsets, we use two indirect but relevant indicators. Quantitatively, the~improvement in mAP@0.75 indicates better localization tightness. Qualitatively, the~feature-response maps in {Figure~\ref{fig:heatmap_comparison}} 
show that PNAFusion produces more concentrated activations around object boundaries and reduces ghosting artifacts compared with the~baselines. 

\begin{table}[H]
	\footnotesize
	\centering
	\caption{Incremental ablation study of the proposed components under the optimal configuration ($L=2$, $k=5$, $N=3$).}
	\label{tab:component_ablation}
	\begin{tabular}{l c c c c c c c c c c c c c}
		\toprule
		\multirow{2}{*}{\textbf{Method}} 
		& \multicolumn{3}{c}{\textbf{mAP@0.5}} 
		& \multicolumn{3}{c}{\textbf{mAP@0.75}} 
		& \multicolumn{3}{c}{\textbf{mAP}} 
		& \multirow{2}{*}{\textbf{Params (M)}} \\
		\cmidrule(lr){2-4} \cmidrule(lr){5-7} \cmidrule(lr){8-10}
		& \textbf{Person} & \textbf{Car} & \textbf{Bicycle} 
		& \textbf{Person} & \textbf{Car} & \textbf{Bicycle} 
		& \textbf{Person} & \textbf{Car} & \textbf{Bicycle} & \\
		\midrule
		NiNfusion & 85.8 & 89.8 & 64.0 & 31.7 & 63.5 & 8.5 & 40.2 & 58.3 & 22.1 & 75.4 \\
		ICAFusion & 84.9 & 89.8 & 73.8 & 28.6 & 60.2 & 12.6 & 38.4 & 55.2 & 28.4 & 120.2 \\
		+PNDCA & 84.8 & 89.8 & 75.3 & 29.9 & 60.7 & 8.9 & 39.1 & 55.3 & 26.2 & 130.9 \\
		+ IterFeed & 87.2 & 91.1 & 74.3 & 38.4 & 65.6 & 14.2 & 43.4 & 59.4 & 28.8 & 152.9 \\
		\bottomrule
	\end{tabular}
\end{table}

The results show that when only the simple NiN Fusion strategy is used, the~model achieves an mAP of only 40.2, indicating that direct linear feature stacking is insufficient to effectively model the complex nonlinear correspondence between multispectral images. After~introducing the proposed PNDCA mechanism, the~overall mAP does not improve significantly. However, once the iterative feedback mechanism is further incorporated on top of this design, the~model performance reaches the peak value of~43.8.

Notably, the~gain in the high-precision detection metric mAP$_{75}$ is particularly significant, increasing from 33.8 to 39.4. This strongly demonstrates that, through recurrent feedback of differential features, the~model can progressively refine the alignment errors produced in previous rounds. In~each iteration, the~offset estimated in the preceding round serves as a prior, enabling the features to gradually converge in the spatial domain. As~a result, the~model exhibits strong robustness when handling edge details and small objects with substantial scale~variation.

Overall, the~PNDCA module and the iterative feedback mechanism exhibit strong synergy in feature alignment and deep cross-modal interaction. Compared with the original baseline, PNAFusion achieves a 3.6 improvement in overall mAP without introducing an excessive computational burden. These results validate the effectiveness of the proposed design philosophy, namely, linear-complexity local interaction and progressive offset refinement, for~multispectral object~detection.

\subsection{Inference Time and Memory~Consumption}
To complement the theoretical complexity analysis, we evaluate the practical efficiency of NiNFusion, ICAFusion, and~PNAFusion on the FLIR test set. As~shown in Table~\ref{tab:efficiency}, PNAFusion requires 156.4G FLOPs, which is lower than the 194.8G FLOPs of ICAFusion because PNCA restricts cross-modal attention to local neighborhoods. PNAFusion also reduces allocated GPU memory from 775.09 MB to 519.61 MB, corresponding to a 33.0\% reduction compared with ICAFusion. This memory reduction is useful for high-resolution multispectral detection on memory-constrained platforms, where memory capacity and bandwidth can be limiting factors. However, PNAFusion does not reduce absolute inference latency. On~the RTX 4090D GPU, its latency is 36.80 ms per image, which is higher than ICAFusion at 25.67 ms and NiNFusion at 17.77 ms. This latency increase mainly comes from deformable bilinear sampling in ADA and repeated refinement in the iterative feedback mechanism. Such operations involve irregular memory access patterns and are less hardware-friendly than standard dense matrix operations. On~the Jetson Orin NX platform, PNAFusion obtains 81.30 ms per image, corresponding to approximately \mbox{12.30 FPS,} while ICAFusion obtains 62.50 ms per image. Therefore, the~practical advantage of PNAFusion should be interpreted as an accuracy--memory/FLOPs trade-off rather than an absolute speed advantage. The~proposed method trades additional latency for stronger local alignment capability, lower theoretical FLOPs than global cross-attention, and~substantially lower allocated GPU~memory. 

\begin{table}[H] 
	\footnotesize
	\centering
	\caption{Inference cost, memory consumption, and latency comparison on the FLIR test set. FPS is computed as $1000/\mathrm{latency}$. The RTX 4090D results are measured with batch size 1 and input resolution $640\times640$. The Orin NX results are measured on an NVIDIA Jetson Orin NX edge platform under the same input resolution. The table is intended to compare the accuracy--memory/FLOPs trade-off rather than to claim lower absolute latency.} 
	\label{tab:efficiency}
	\begin{tabular}{l c c c c c c c} 
		\toprule 
		\textbf{Method} & \textbf{Params (M)} & \textbf{FLOPs (G)} & \textbf{4090D (ms)} & \textbf{FPS 4090D} & \textbf{Orin NX (ms)} & \textbf{Orin NX (FPS)} & \textbf{GPU Memory (MB)} \\ 
		\midrule 
		NiNFusion & 75.4 & 112.5 & 17.77 & 56.27 & 42.64 & 23.45 & 535.85 \\ 
		ICAFusion & 120.2 & 194.8 & 25.67 & 38.96 & 62.50 & 16.00 & 775.09 \\ 
		PNAFusion & 130.9 & 156.4 & 36.80 & 27.17 & 81.30 & 12.30 & 519.61 \\ 
		\bottomrule 
	\end{tabular}
\end{table}

\subsection{Qualitative~Analysis}

Because the public FLIR, M3FD, and~DroneVehicle benchmarks do not provide official image-level illumination labels such as daytime and nighttime, we do not manually split the test sets into day and night subsets. A~manual split would introduce subjective bias, especially for ambiguous scenes such as dusk, dawn, shadows, glare, and~mixed illumination. Instead, we provide qualitative examples covering representative challenging illumination conditions, including low-light nighttime scenes, strong glare, and~cluttered backgrounds. These examples are used to visually examine whether multispectral fusion improves detection robustness under illumination~variation. 

\begin{figure}[H]
	\includegraphics[width=0.99\textwidth]{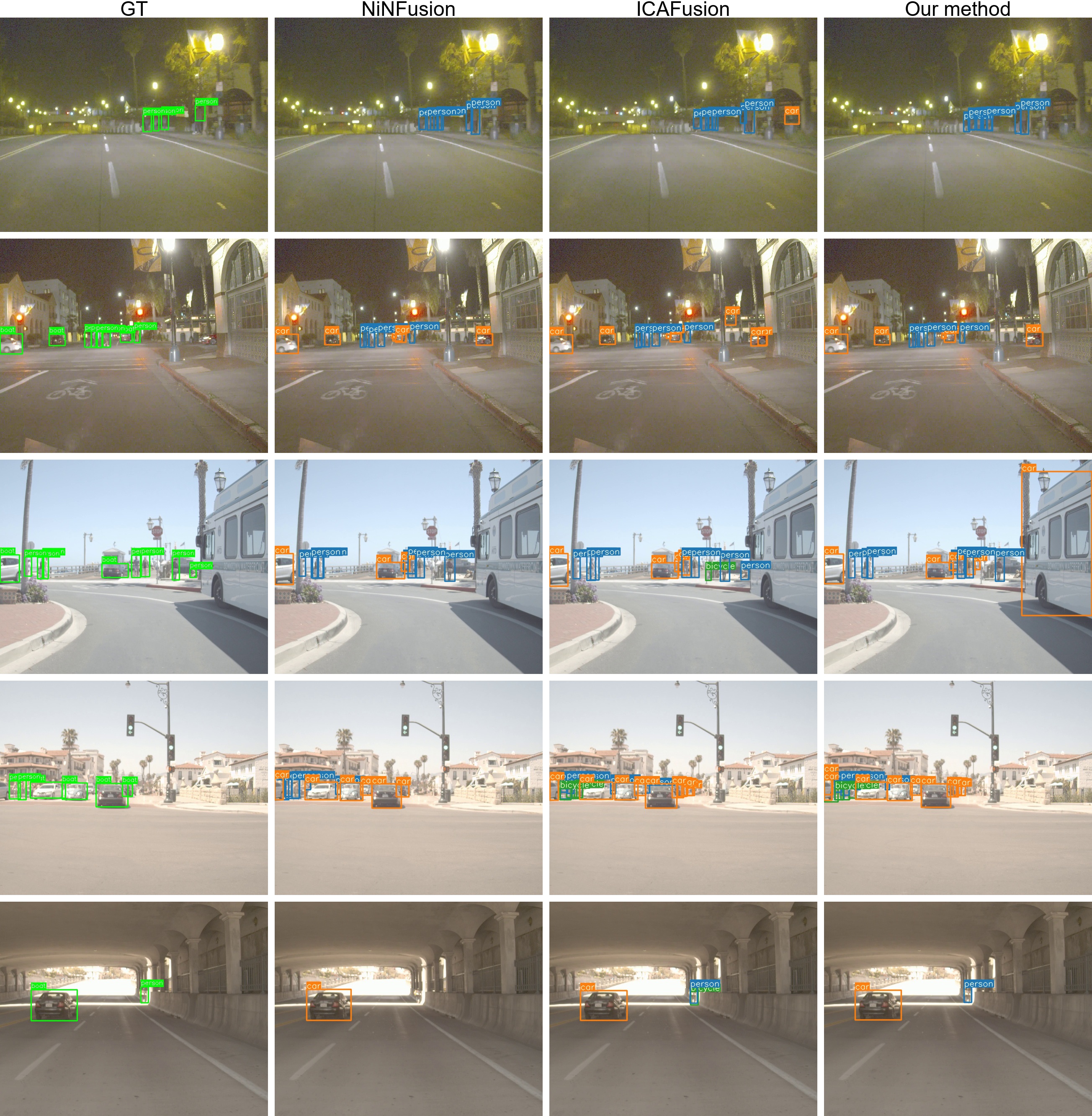}
	\caption{{Qualitative} 
		{comparison} 
		{of} 
		detection results on the FLIR dataset. From~left to right, the~columns show ground truth, NiNFusion, ICAFusion, and~PNAFusion. The~selected examples include low-light, glare, and~cluttered-background scenes. Compared with the baselines, PNAFusion produces tighter bounding boxes and fewer missed detections, indicating that neighborhood-constrained deformable cross-modal fusion improves robustness under challenging illumination and weak alignment~conditions.} 
	\label{fig:qualitative_comparison}
\end{figure}

Figure~\ref{fig:qualitative_comparison} presents qualitative comparisons on representative FLIR scenes, including nighttime roads, urban intersections, strong illumination transitions, and~partially occluded traffic targets. We compare the proposed method with NiNFusion and ICAFusion, and~visualize the predictions together with the ground truth. Overall, our method produces detection results that are closer to the ground truth in both object localization and category prediction, especially in challenging cases with low illumination, weak cross-modal misalignment, and~cluttered~backgrounds.

In nighttime scenes (the first two rows in Figure~\ref{fig:qualitative_comparison}), small pedestrians are easily overwhelmed by glare, low contrast, and~background noise. The~competing methods either miss distant pedestrians or introduce extra false positives around bright street lamps and roadside structures. In~contrast, our method detects more complete pedestrian instances with tighter bounding boxes. This result suggests that the proposed deformable alignment module helps establish more reliable correspondence between visible and thermal features before fusion, which is particularly beneficial when the two modalities are not perfectly~aligned.

The third and fourth rows in Figure~\ref{fig:qualitative_comparison} further show that our method is more robust in crowded street scenes containing multiple vehicles and overlapping objects. NiNfusion and ICAFusion tend to generate redundant detections or inaccurate object extents, especially near large foreground vehicles and high-response background regions. By~restricting cross-modal interaction to local pixel neighborhoods, our method suppresses irrelevant long-range interference and yields cleaner predictions with fewer obvious false alarms. This behavior is consistent with the quantitative improvements in localization-sensitive~metrics.

The last row in Figure~\ref{fig:qualitative_comparison} highlights a tunnel scene with strong brightness transition, where the pedestrian target is small and located near the image boundary. NiNfusion fails to recover this target, while ICAFusion introduces category confusion. Our method still identifies both the vehicle and the pedestrian with more accurate localization. This example indicates that iterative refinement is useful for progressively correcting cross-modal inconsistency and preserving subtle target cues under extreme illumination~changes.

\begin{figure}[H]
	\includegraphics[width=0.98\textwidth]{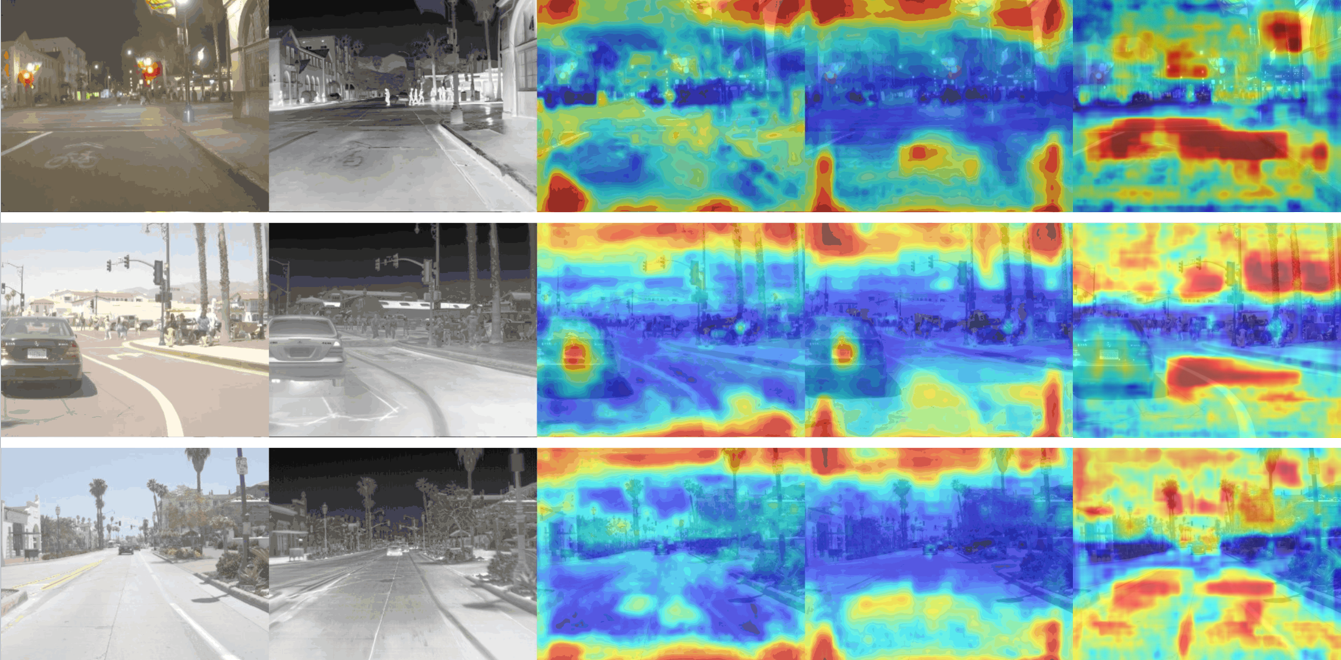}
	\caption{{Feature-response} 
		{visualization} 
		for different fusion strategies. From~left to right:
		RGB image, infrared image, NiNFusion, ICAFusion, and~PNAFusion. Baseline methods without adaptive deformable alignment tend to produce diffuse responses, boundary ghosting, or~multiple disconnected activation peaks around weakly aligned objects. In~contrast, PNAFusion generates more concentrated activations near object regions and suppresses irrelevant background responses. This visualization provides qualitative evidence that the learned local offsets in ADA and the neighborhood-constrained aggregation in PNCA help improve cross-modal spatial consistency.}
	\label{fig:heatmap_comparison}
\end{figure}

Figure~\ref{fig:heatmap_comparison} further visualizes the cross-modal response distributions of different methods. Compared with NiNFusion and ICAFusion, PNAFusion exhibits more concentrated activations around pedestrians and vehicles while suppressing diffuse responses on roads, sky regions, and~other irrelevant background structures. This behavior is particularly evident in the red boxed regions, where the proposed method preserves salient target cues under cluttered illumination and weak cross-modal misalignment, providing intuitive evidence for the effectiveness of the ADA and PNCA~modules.

Overall, the~qualitative results confirm the main advantage of PNAFusion: it improves cross-modal detection reliability not only on standard scenes but also under difficult conditions where precise alignment and local interaction are essential. These visual comparisons provide intuitive evidence that the proposed adaptive deformable alignment and pixel-neighborhood cross-attention jointly enhance detection completeness, localization quality, and~resistance to background~distraction.

\subsection{Failure Case and Limitation~Analysis}

\begin{figure}[H]
	\includegraphics[width=0.98\textwidth]{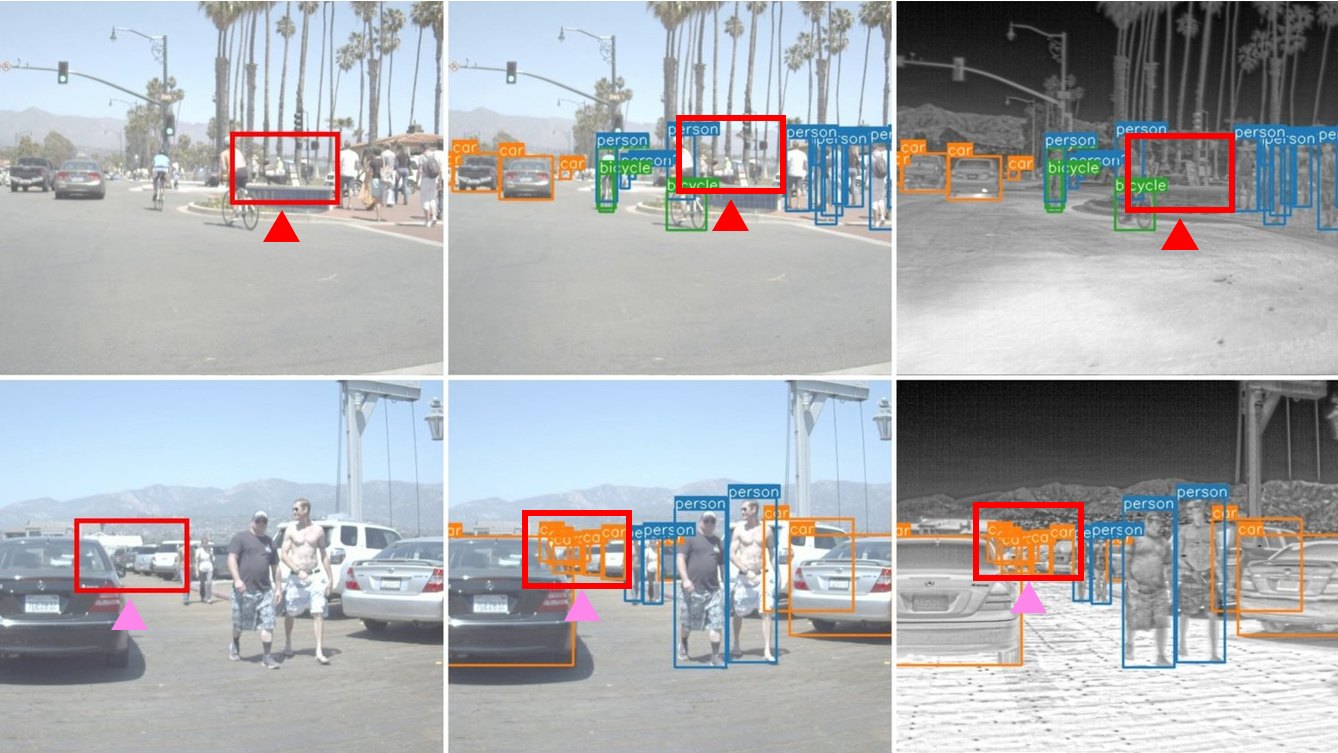}
	\caption{{Representative} 
		{limitation} 
		cases of PNAFusion. From~left to right in each group: original image annotated with failure regions, RGB image, and~infrared image. The~pink triangle markers indicate false detections of the baseline, while the red triangle markers indicate missed detections. In~the first scene, extremely small pedestrian targets may still be missed because their visible texture is weak and their thermal response occupies only a few pixels. In~the second scene, heavily overlapped vehicles may lead to duplicate or inaccurate detections because adjacent instances produce highly similar local cross-modal~responses.}
	\label{fig:limitation_cases}
\end{figure}

Although PNAFusion improves cross-modal alignment and local feature interaction, it may still fail in several challenging scenarios. Figure~\ref{fig:limitation_cases} presents representative failure cases. In~each group, the~columns from left to right show the original image annotated with failure regions, the~RGB image, and~the infrared image. The~pink triangle markers indicate false detections of the baseline, while the red triangle markers indicate missed~detections.

As indicated by the red triangle markers in the first scene of Figure~\ref{fig:limitation_cases}, when pedestrian targets are extremely small and located in a complex traffic background, the~proposed model may still produce missed detections. This failure is mainly caused by the limited spatial resolution of deep feature maps and the weak target evidence in both modalities. Although~the thermal modality provides complementary cues under low illumination, very small pedestrians may occupy only a few pixels after feature downsampling. In~this case, the~local neighborhood used by PNCA may contain insufficient discriminative information, and~the learned deformable offsets in ADA cannot recover target details that have already been weakened by the~backbone.

The second scene, highlighted by the pink triangle markers, shows another limitation under severe object overlap. When multiple vehicles are spatially close or partially occluded, their RGB contours and infrared responses may merge into a compact high-response region. Under~this condition, PNAFusion can still enhance local cross-modal consistency, but~the neighborhood-constrained interaction may aggregate similar features from adjacent instances. As~a result, the~detector may generate duplicate boxes, overlapping detections, or~false positives around the same vehicle group. This indicates that local alignment alone cannot fully resolve instance-level ambiguity when object boundaries are heavily~occluded.

These failure cases suggest that PNAFusion is most reliable when complementary modal cues are locally consistent and object instances remain sufficiently separable. Its performance may degrade for extremely small targets, severe occlusion, dense overlapping objects, and~scenes where both RGB and infrared modalities provide weak or ambiguous evidence. Future work will investigate higher-resolution feature preservation, scale-aware supervision for tiny objects, and~instance-aware matching constraints to reduce missed detections and duplicate detections in such challenging~scenarios. 

\section{Conclusions}  
\label{sec5}
This paper presented PNAFusion, a~progressive pixel-neighborhood deformable cross-attention framework for multispectral object detection. The~proposed method addresses two key challenges in visible--thermal fusion: weak cross-modal misalignment and the high memory cost of global cross-attention. By~coupling Adaptive Deformable Alignment with Pixel-Neighborhood Cross-Attention, PNAFusion performs content-aware local alignment and restricts cross-modal interaction to semantically relevant neighborhoods. The~iterative feedback mechanism further refines feature correspondence across multiple rounds, improving localization robustness under imperfect registration. Experiments on FLIR, M3FD, and~DroneVehicle demonstrate that PNAFusion achieves highly competitive detection accuracy across different multispectral benchmarks. In~particular, the~improvement in mAP@0.75 and the qualitative feature-response visualizations indicate that the proposed design helps alleviate spatial ghosting and improves boundary localization. Efficiency analysis shows that PNAFusion reduces theoretical FLOPs and allocated GPU memory compared with global cross-attention. However, it also introduces additional inference latency due to deformable bilinear sampling and iterative refinement. Therefore, the~main practical benefit of PNAFusion lies in its accuracy--memory/FLOPs trade-off rather than absolute inference speed. Several limitations remain. First, the~current benchmarks do not provide dense pixel-level cross-modal alignment ground truth, making it difficult to directly measure the absolute accuracy of learned offsets. Second, public datasets do not consistently provide official day/night labels, so illumination-specific quantitative evaluation is not included. Third, deformable sampling is not fully optimized for all hardware platforms and may cause irregular memory access. Future work will investigate hardware-friendly deformable sampling, explicit offset supervision when alignment annotations are available, and~more fine-grained evaluation under different illumination and weather~conditions.


\end{document}